\documentclass[twoside,leqno,twocolumn]{article}
\usepackage{ltexpprt}
\usepackage{amssymb}
\usepackage{graphicx}

\begin{document}

\title{Early text classification: a Na\"ive solution\thanks{Manuscript submitted to SIAM International Conference on Data Mining 2016. This work was supported by CONACyT grants No. CB-2014-241306 (\emph{Clasificaci\'on y Recuperaci\'on de Im\'agenes Mediante T\'ecnicas de Miner\'ia de Textos}) and PN-2015 (\emph{Caracterizaci\'on de usuarios en redes sociales: hacia un enfoque multimodal y multidominio}).  }}
\author{Hugo Jair Escalante\thanks{Computer Science Department, Instituto Nacional de Astrof\'isica, \'Optica y Electr\'onica, 72840, Puebla, M\'exico} \\
\and
Manuel Montes$^\dag$
\and
Luis Villase\~nor$^\dag$
\and
Marcelo L. Errecalde\thanks{Computer Science Department, Universidad Nacional de San Luis, D5700HHW, Argentina}
}
\date{}

\maketitle

%\pagenumbering{arabic}
%\setcounter{page}{1}%Leave this line commented out.

\begin{abstract} \small\baselineskip=9pt Text classification is a widely studied problem, and it can be considered \emph{solved} for some domains and under certain circumstances. There are scenarios, however, that have received little or no attention at all, despite its relevance and applicability. One of such scenarios is early text classification, where one needs to know the category of a document by using partial information only. A document is processed as a sequence of terms, and the goal is to devise a method that can make predictions as fast as possible. The importance of this variant of the text classification problem is evident in domains like sexual predator detection, where one wants to identify an offender as early as possible. This paper analyzes the suitability of the standard na\"ive Bayes classifier for approaching this problem. Specifically, we assess its performance when classifying documents after seeing  an increasingly number of terms.  A simple modification to the standard na\"ive Bayes implementation allows us to make predictions with partial information. To the best of our knowledge Na\"ive Bayes has not been used for this purpose before. Throughout an extensive experimental evaluation we show the effectiveness of the classifier for early text classification. %We show the generality of the approach by reporting results in early gesture recognition as well.
What is more, we show that this simple solution is very competitive when compared with state of the art methodologies that are more elaborated. We foresee our work will pave the way for the development of more effective early text classification techniques based in the na\"ive Bayes formulation.
\end{abstract}


\section{Introduction}

Text classification is the task of assigning documents to its correct categories~\cite{sebastiani}. This is  one of the most studied topics within natural language processing. Advances in the last two decades have made significant progress and nowadays the text classification problem is considered to be solved in some scenarios and under certain circumstances (e.g., news classification with plenty of data). There are, however, settings of the text classification problem that have received little attention despite the wide applicability they may have. One of such scenarios is that of \emph{early text classification}, which  deals with the development of predictive models that are capable of determining the class a document belongs to as soon as possible. A text is assumed to be processed sequentially, starting at the beginning of the document and reading input words one by one. It is desired to make predictions with as low information as possible.

The early text classification topic has received little attention in the community, and there exist only a few works that have approached similar scenarios~\cite{sequentialTC} (please note that in this work the problem is not stated as one of early recognition). Despite its low popularity, this topic has a major potential in practical applications. For instance, consider the problem of detecting sexual predators in chat conversations. Here, the goal is to sequentially read a conversation and to determine as fast as possible whenever a sexual predator is involved; clearly, a detection using the whole conversation can only be used for forensics rather than for prevention. Other sample applications include, any kind of conversation analysis that requires of a fast response, (e.g., cyber-bullying prevention, adaptive/intelligent answering systems); trending-topic discovery (e.g., analyzing comments on social networks and determining as soon as possible whenever a topic will become a trend); content filtering (e.g., filtering inappropriate/ilegal content in local networks), author profiling (e.g., knowing the age, gender or interest of a person by using as few written information as possible) etcetera. %improving adaptability on interactive systems using text as input (e.g., chat support-online support systems), etcetera.

This paper explores the suitability of one of the most popular methods for text classification, i.e., na\"ive Bayes~\cite{mcallum,sebastiani}, to approach the early-classification setting: \emph{early na\"ive Bayes}. Specifically, we evaluate the capabilities of this classifier to make predictions when \emph{seeing} an increasing number of terms from documents. %we assess the performance of na\"ive Bayes in the early text classification setting
 A simple modification to the standard na\"ive Bayes implementation allows us to make predictions with partial information.  %This is not a novel discovery but an adaptation of the original implementation.
Despite its simplicity, the proposed extension obtains competitive performance in standard text classification tasks and in sexual predator detection. In fact we show that the proposed modification compares favorably with the only existing work that addresses a similar task. Hopefully, our work will motivate research on further extensions to this classifier for early text classification. %Therefore, our work can be considered a \emph{maximum improvement with minimum innovation} solution to the early classification problem, that can be improved in a number of ways and, hopefully, our work will motivate further research in this topic.

The remainder of this  %rest of this
paper is organized as follows. Next section reviews related work on early text classification and on extensions to na\"ive Bayes to face closely related problems. Then, Section~\ref{sec:NB} describes na\"ive Bayes classifier and the modification we propose to make early predictions. %Next
Section~\ref{sec:experiments} reports experimental results that show the effectiveness of the proposal. Section~\ref{sec:conclusions} presents conclusions and discusses future work directions.

\section{Related work}
\label{sec:rw}
This section reviews related work on both: early text classification and extensions to na\"ive Bayes to face similar problems.

\subsection{Early text classification}

To the best of our knowledge, the early text categorization problem has been approached only in~\cite{sequentialTC}; although the authors' main focus was not on making predictions earlier but on improving the classification performance with a \emph{sequential reading approach}.  In that work, the authors process documents in a sentence-level basis. Every time $t$, the authors read a sentence and attempt to determine the class of the document, where multi-label classification is allowed. They proposed a Markov decision process (MDP) to approach the problem, where two possible actions were allowed: read next sentence, or classify. Each sentence has to be represented by its \emph{tfidf} representation and a classifier is trained to learn good/bad state-action pairs (10,000 examples were randomly generated) on a high-dimensional space.

The performance of their method was evaluated in standard text classification data sets. Although the performance of such method is competitive (it was compared to a SVM classifier), it remains unknown whether a much more simpler approach would be as effective as the complex procedure in~\cite{sequentialTC}. In Section~\ref{sec:experiments} we compare the proposed extension of na\"ive Bayes with the previous work. We show our proposal is competitive in terms of performance, but also has the following advantages: it is scalable in the number of categories (the MDP evaluated every possible state after reading each sentence, ours simply adds probabilities); it is able to make predictions with as low information as no-word (using priors-only information, but the most important aspect is that it can make predictions at anytime); it process documents in a word-level basis (i.e., one word added at a time, while the MDP requires processing whole sentences); training is much more efficient (same training complexity as an standard na\"ive Bayes classifier, the MDP requires of high-complexity training procedures) and the resultant model is way more simple.

Although the early text classification problem has not been studied elsewhere, it is worth mentioning works that have approached related tasks. In~\cite{hmmzaragoza}, the authors propose a hidden Markov model (HMM) to classify passages within documents. The task is information retrieval and a document is considered as relevant or irrelevant (i.e. two classes) to a given category/query. The document is decomposed into passages, each of which is considered by the HMM as relevant or irrelevant to the classification. No attempt is made to perform classification early, although it is interesting that the proposed model is a generalization of the multinomial na\"ive Bayes we consider in this work (again, for the two-class whole-document classification problem). %Documents can belong to one of two classes (relevant or irrelevant to a catergory/query) and passages are modeled as
 %The proposed method processed documents on a passage-based level, where the goal was to classify a passage as relevant or irrelevant.
%Stefan et al.~\shortcite{recurrent} proposed a recurrent neural network to process documents sequentially.

In~\cite{datumwise} the authors extend the MDP proposed for sequential text classification to deal with any other type of data. The formulation is almost the same as in~\cite{sequentialTC}, although this time the MDP can decide what feature to sample from the instance under analysis (i.e., there is no sequential input). Furthermore, the MDP is equipped with a mechanism that aims to minimize the number of features to use for classification. Clearly, this extended MDP is not applicable to the early text classification domain (words cannot be chosen from documents, they appear sequentially).

Summarizing, it is remarkable the little attention that early text classification has received so far, this may be due to the fact that not so many applications in the past required to cope with this problem. Nowadays, however, the \emph{online status} of the world population, requires of technology that can anticipate the prediction of certain events with the goal of preventing undesired effects or, on the other hand, to act as fast as possible to take the leadership on information technology.

\subsection{Extending na\"ive Bayes}

Na\"ive Bayes has been used extensively in text mining and within machine learning in general, because of its high performance in several domains, several modifications and extensions have been proposed to augment the scope of the classifier. Related to our work, the following extensions have been reported in the literature:
\begin{itemize}
\item \textbf{Alleviating independence assumption of Na\"ive Bayes.} This is perhaps the most studied topic in terms of extending the mentioned classifier. The independence assumption may be too strong for some domains/applications, therefore, several works have been proposed that try to relax it. Most notably TAN~\cite{tan}, AODE~\cite{aode}, and WANBIA~\cite{alleviating} %AODE~\cite{aode}
    extensions
    have reported outstanding results. Nevertheless, the focus here is on relaxing the attribute independence assumption, and not on working with partial information. One should note, however, that this extended versions of na\"ive Bayes can be well suited for early text classification, as attribute-dependency information  can help the algorithm to classify texts earlier.

\item \textbf{Anytime na\"ive Bayes.} The goal of this type of extensions is to provide na\"ive Bayes with mechanisms that allow it to make predictions at anytime~\cite{anytime1,anytime2}. This means that the algorithm has to be ready to provide a prediction under time constraints: the classifier can spent increasing amounts of time for doing inference, but it must provide an answer when requested; usually accuracy increases as more time is allowed. This type of methods is  related to our proposal in that the system has to be ready to make predictions at anytime, however, the granularity of information processing is different: in anytime classification a whole instance is seen, whereas in early text classification, part of an instance is available.

\item \textbf{Incremental na\"ive Bayes.} Refers to developing learning and inference mechanisms to allow the classifier be trained in an online learning setting~\cite{incrementaliberamia,evolvingNB}. That is, reading a sample (or batch of samples at a time), the model makes predictions for the incoming samples and then it is provided with the correct labels, next, model parameters have to be updated accordingly. This type of methods are related to our proposal in that partial information is processed incrementally, although one should note that information units are instances and not words/attributes. %Anyway, research in this subfield can be helpful to early classification because
\item \textbf{Na\"ive Bayes for incomplete information.} These extensions aim at helping na\"ive Bayes to deal with missing information, usually, at the attribute level. For instance by equipping the classifiers with mechanisms to work under highly-sparse representations (e.g., in short text categorization)~\cite{Shen:2009:ETR:1645953.1646192,cabrera,smoothing,otroNB}. These methods are mostly based on smoothing attribute-class probabilities and often use co-occurrence statistics. Although not dealing with early text classification, this type of methods are relevant because smoothing plays a key role when working with partial information (everything not seen so far has to be smoothed).
\end{itemize}

Summarizing, there have been many attempts to improve and extend na\"ive Bayes to be robust against several limitations, however, to the best of our knowledge,
%As it can be seen from the review above,
%From the above review, we can say that,  to the best of our knowledge, the na\"ive Bayes classifier
it has not been used for early text classification before. This is somewhat surprising given that, as shown in the next section, the na\"ive Bayes classifiers can naturally deal with partial information.

\section{Early text classification with Na\"ive Bayes}
\label{sec:NB}

This section describes the way we use na\"ive Bayes classifier for early text classification. %In the following we describe the classifier in terms of text classification however, the same reasoning

\subsection{Na\"ive Bayes classifier}

We first describe the standard na\"ive Bayes classifier. Consider a data set: $\mathcal{D} = (\textbf{x}_i, y_i)_{\{1, \ldots, N\}}$ with $N$ pairs of instances ($\textbf{x}_i$) and labels ($y_i$) associated to a supervised classification problem. Assuming that $\textbf{x}_i\in \mathbb{R}^q$ and $y_i \in C = \{1, \ldots, K\}$ we have  a $K-$class classification problem with numeric\footnote{One should note that in text classification we can transform any document to a numeric vector with the bag of words representation, i.e., a vector of length $q$, where $q$ is the vocabulary size and each element of the vector indicates the relevance of a term for describing the content of the document.} attributes.
%; and considering $g_k(\textbf{x}_i) \in [-1, 1]$ as the classifier output $g_k$ for instance $\textbf{x}_i$.
%$g_k$ represents the predictor's confidence value for $\textbf{x}_i$ class. Every $g_k$ term can be modeled as a function $g_k : \mathbb{R}^d \rightarrow [-1, 1]$, where the predicted class for $\textbf{x}_i$, defined by $\hat{y}_i$, is obtained as follows: $\hat{y}_i = sign (g_k(\textbf{x}_i))$.

Under the na\"ive Bayes classifier, the class for an unseen instance $\textbf{x}_T =\langle x_{T,1}, \ldots, x_{T,q} \rangle$ is given by:
\begin{equation}\label{eq:NBD}
\hat{C} = \arg\max_{C_i} P(C_i | \textbf{x}_T) %\sum_{i=1}^n ||  d_i - NN_C(d_i)||
\end{equation}

From Bayes' theorem it follows that the posterior probability above can be estimated as:
\begin{equation}\label{eq:NBP1}
P(C_i | \textbf{x}_T) = \frac{P(\textbf{x}_T | C_i ) P(C_i)}{P(\textbf{x}_T)}
\end{equation}

The denominator can be removed from Equation (\ref{eq:NBD}) as it does not affect the decision: %\footnote{Alternatively, one can take the marginal: $P(\textbf{x}_T) = \sum_i P(\textbf{x}_T | C_i) P(C_i)$.} we have:
\begin{equation}\label{eq:NBP2}
P(C_i | \textbf{x}_T) \approx P(\textbf{x}_T | C_i ) P(C_i)
\end{equation}

The assumption of na\"ive Bayes is that the probability of  occurrence of attributes of $\textbf{x}_T$ is independent given its class, that is:
\begin{equation}\label{eq:NBP25}
P(C_i | \textbf{x}_T) \approx \prod_{j=1}^{q} P(x_{T,j} | C_i ) P(C_i)
\end{equation}

The maximum likelihood estimation for the prior of class $C_i$ is  given by:
\begin{equation}\label{eq:prior}
\hat P(C_i) = \frac{|X_i|}{N}
\end{equation}
where $X_i$ is the set of all instances in $\mathcal{D}$ that are labeled with class $C_i$. Hence, the key of the na\"ive Bayes classifier lies in the estimation of $P(\textbf{x}_T | C_i)$, or more precisely of $\prod_{j=1}^{q} P(x_{T,j} | C_i )$. Depending on the type of data (e.g., binary, discrete, or real) a different distribution may be assumed for computing $P(x_{T,j} | C_i )$ (e.g., Bernoulli, Multinomial, or Gaussian, respectively). In text classification one of the most effective implementations is based in the multinomial distribution, when documents are represented by its term-frequency representation (i.e., we know for each document, the number of times each term from the vocabulary occurs)~\cite{mcallum,multinomial}. Accordingly, we focus in this implementation, this means we assume w.l.o.g.: $\textbf{x}_i\in \mathbb{Z}_+^q$ (i.e. the representation of a document is a vector of frequency values / integers).
%Assuming a Bernoulli distribution we have:
%\begin{equation}\label{eq:NBP3}
%P(C_i | \textbf{x}_T) \approx P(\textbf{x}_T | C_i ) P(C_i)
%\end{equation}

Assuming a multinomial distribution for the model we have that the maximum likelihood estimation for the term of interest is:
\begin{equation}\label{eq:NBP4}
P(\textbf{x}_T | C_i ) \approx \prod_{j=1}^{q} \hat P(x_{T,j} | C_i )^{f_{j,T}}
\end{equation}
where $f_{j,T}$ is the value of the $j^{th}$ attribute in instance $\mathbf{x}_T$ (in text classification $f_{j,T}$ is  the frequency of occurrence of the $j^{th}$ term in document $T$), and
\begin{equation}\label{eq:NBP5}
\hat P(x_{T,j} | C_i ) = \frac{1 + F_{j,C_i}}{q + \sum^q_k{F_{k,C_i}}}
\end{equation}
where $F_{l,C_i}$ is the sum of values of the $l^{th}$ attribute in documents of class $C_i$. The derivation from Equation (\ref{eq:NBP4}) removes factorial terms that do not affect the final decision. %, and $F_{k,C_i}$ is the value of attribute
%\begin{equation}\label{eq:NBP5}
%P(C_i | \textbf{x}_T) \approx \prod_{j=1}^{N_T} P(x_j | C_i ) P(C_i)
%\end{equation}
For more details we refer the reader to~\cite{mcallum,multinomial}.
In the description above we did not assume a text categorization problem because the same results apply to any type of (multinomial-distributed) attributes. In the following we use text-mining terminology, but we emphasize the description is generalizable to other problems.

\subsection{Early Na\"ive Bayes}

%Assuming a maximum likelihood estimator for the prior $P(C_i)$, one should note that the estimation of $P(C_i | \textbf{x}_T)$ depends only on the product of term probabilities conditioned on the classes $P(x_j | C_i )$.
In early text classification we assume that during training we have full documents, therefore, the same training procedure as the standard na\"ive Bayes classifier is performed for estimating the necessary probabilities\footnote{One may also train na\"ive Bayes with partial documents, however, in that case the probability estimates associated to the model are not reliable because they are obtained from reduced documents. In preliminary experiments we corroborated this fact.  }. The difference comes at inference time: when classifying a new document we assume we read it in sequential order starting from the beginning  (i.e. the first word from top to bottom and from left to right). W.l.o.g.\footnote{One should note that we can take steps of any length, instead of processing word-by-word. }, at time $t$ we assume we have read the first $t-$terms in the document (i.e., one word is read at each time). Let $d_T$ denote the document we want to classify, where it contains $M_{d_T}$ words, then, $d_T = w_1, w_2, \ldots, w_{M_{d_T}}$.

We notice from Equations (\ref{eq:prior}-\ref{eq:NBP5}) that in fact we can make predictions for document $d_T$ regardless the amount of information we have read from it: at time $t$ we know that $d_T = w_1, \ldots, w_t$, therefore, we can generate a bag-of-words $\mathbf{x}_T$ representation for $d_T$ as follows $\mathbf{x}_T = \langle \mathbf{x}_{T,1}, \ldots, \mathbf{x}_{T,q} \rangle$, where $\mathbf{x}_{T,j}$ indicates the frequency of occurrence of the $j^{th}$ term in document $d_T$ (i.e., a \emph{tf} weighting scheme).  Terms not occurring the $d_T$ or not seen so far at time $t$ are assigned values of $\mathbf{x}_{T,j}=0$. With this representation we can use Equation (\ref{eq:NBP2}) directly to classify the document. Actually, we can attempt to classify document $d_T$ without having read any information! (i.e., with $t=0$), of course the probability will be dominated by the priors, see Equation (\ref{eq:prior}). Simply as this, we can use na\"ive Bayes to perform early classification.

%\subsubsection{Analysis}
We now briefly analyze what are the main components in play when making predictions early. At time $t$ one can rewrite Equation (\ref{eq:NBP25}) as:
\begin{equation}\label{eq:NBP6}
\footnotesize{
P(C_i | \textbf{x}_T) \approx P(C_i)  \prod_{j: j\in d_T} P(x_{T,j} | C_i )  \\ \prod_{k: k \not\in d_T} P(x_{T,k} | C_i )}
\end{equation}
the second product (over $j\in d_T$) accounts for the terms appearing in the document (probabilities are affected by the frequency of occurrence of such terms in $d_T$ so far); %  and their probability depends on the frequency of occurrence of terms in the document,
the third product (on $k \not\in d_T$) simply reduces to 1 (because of the exponent in Equation (\ref{eq:NBP4})). %multiply the smoothed probability for terms not occurring in $d_T$ or not seen at time $t$, that is: $\hat P(x_{T,k} | C_i ) = \frac{1}{q + \sum^q_k{F_{k,C_i}}}$ for these terms.
Therefore, for small values of $t$, the priors dominate the decision, as $t$ increases the content of the document will dominate the other products. Therefore, the way these three components are estimated can be crucial for improving the performance of na\"ive Bayes in early classification. %For instance, adaptive priors and non-zero probability estimates for unseen terms are promising venues for research on early classification with na\"ive Bayes. %, we are doing research in this direction in our group. %this is work in progress in our group.

Despite the simplicity of this early text classification approach, we will see in the next section that it compares favorably with a more complicated solution from the state of the art. We show its validity in a variety of problems. This paper motivates further work on extending this model for early text classification.  For instance, one can define/modify adaptive priors that change as the value of $t$ increases; we can implement the same idea with methods that take into account term-dependencies (see e.g.,~\cite{tan,aode,alleviating}) in order to increase the predictive power of the classifier; also one can adopt advanced/alternative smoothing techniques to account for partial and missing information properly~\cite{Shen:2009:ETR:1645953.1646192,cabrera,smoothing}; as well as many other possibilities. The main goal of this paper is to show that na\"ive Bayes can be used for early text classification and that its performance is competitive with the single existing solution to this problem. We foresee our work will pave the way for development of a new type of models.

\section{Experiments and results}
\label{sec:experiments}

For experimentation we considered the data sets described in Table~\ref{tab:datasets}. We considered three standard thematic text categorization tasks (also used in~\cite{sequentialTC}) and a data set for sexual predator detection~\cite{sexpred}. % and a gesture recognition benchmark~\cite{lap2014}, 
All of the data and our code will be made available under request for future comparisons. In the subsections below we provide details on each data set and report the corresponding experimental results obtained  with them.
\begin{table}[htb]
%\begin{center}
\centering
\scriptsize{
\begin{tabular}{|c|c|c|c|c|c|}
\hline
\multicolumn{6}{|c|}{\textbf{Text categorization}}\\\hline
\textbf{Data set}&\textbf{Classes}&\textbf{Terms}&\textbf{Red.V.}&\textbf{Train}&\textbf{Test}\\\hline
Reuters-8&8&23583&2483&5339&2333\\ %2483
%Reuters-10&10&25283&6287&2811\\
20-Newsgroup&20&61188&6894&11269&7505\\ %6894
%TDT-2&30&36771&6576&2818\\
WebKB&4&7770&3727&2458&1709\\\hline % 3727
\multicolumn{6}{|c|}{\textbf{Sexual predator detection}}\\\hline
SPD&2&155886&6770&6588&15329\\\hline
%\multicolumn{6}{|c|}{\textbf{Gesture recognition}}\\\hline
%Montalbano&20&2000&2000&6850&3454\\\hline
\end{tabular}}
\caption{Data sets considered for experimentation. Red. V. is the number of terms when a reduced vocabulary is used. }\label{tab:datasets}
%\end{center}
\end{table}

Text data sets were processed as follows: stop words were removed, then stemming was applied, next the bag-of-words representation was obtained using the TMG toolbox, a term-frequency (\emph{tf}) weighting scheme was used~\cite{tmg}. All of the data were processed in Matlab$^R$. For most experiments we used reduced vocabularies, that is, we used only a subset of the most frequent words/terms (see column 4 in Table~\ref{tab:datasets}), we proceeded like this for efficiency, nevertheless we also report results with full-vocabularies in text categorization data sets. %; Table~\ref{tab:datasets} specifies the number of terms considered when reduced vocabularies were used.

%In our implementation we worked with the na\"ive Bayes (multinomial) implementation from WEKA~\cite{WEKA}.
%The na\"ive Bayes classifier was implemented in Matlab and
In addition to the comparison to the state of the art, we considered a linear SVM classifier as baseline, since this is a \emph{mandatory} baseline in text classification~\cite{TCsvm,sebastiani}. SVM was used in early classification similarly as the na\"ive Bayes model: it was trained with complete documents, and for making predictions, the bag of words of a document up to time $t$ is obtained and feeded to the SVM classifier. In preliminary experimentation we compared SVM with \emph{tf} and \emph{tfidf} weighting schemes, we report the performance of SVM with the latter scheme because we obtained better results with this configuration. %We considered two variants of the SVM: using  \emph{tf} and \emph{tfidf} weighting schemes. %\footnote{One should note that even when we can use the SVM for early classification, the model is not expected to perform well with partial information. This is because the vector of weights was learned with full documents and attempts to make predictions for much more feature vectors that are very different from the weights vector (mainly when few information has been read, see below). }. It is important to mention that no parameter optimization was performed for any method.

In all of our experiments we report the performance of the early text classifiers when varying the percentage of the words in test documents (same procedure as in~\cite{sequentialTC}). Macro-average $f_1$ measure was used for multiclass text categorization problems and $f_1$ of the minority class (i.e., predators) for the sexual predator detection data set. 
Ideally, the performance of a good early text classifier should draw a curve close to the $y-axis$ (see figures below): i.e., better performance with less information.  %Please note that in~\cite{sequentialTC} documents were also split in this way.
A different problem, not evaluated in this paper, is that of triggering a prediction whenever the classifier is sure about the class of a document. Please note, however, that simple triggering mechanisms can be derived for our proposed formulation, e.g., after seeing a predefined number of words, or when the difference between the most probable and the second most probable class exceeds a threshold, and so on.
%Anyway, we compare our work with the method in~\cite{sequentialTC}.

%We consider both thematic and non-thematic text-classification problems, in the former problems the goal is to classify documents into thematic categories (e.g., ``sports vs. religion''), whereas in the considered non-thematic tasks the goal is to associate documents to authors (i.e., modeling the writing style of authors).
\subsection{Early text categorization}
\begin{figure}[htb!]
    \centering
    \includegraphics[scale=0.37]{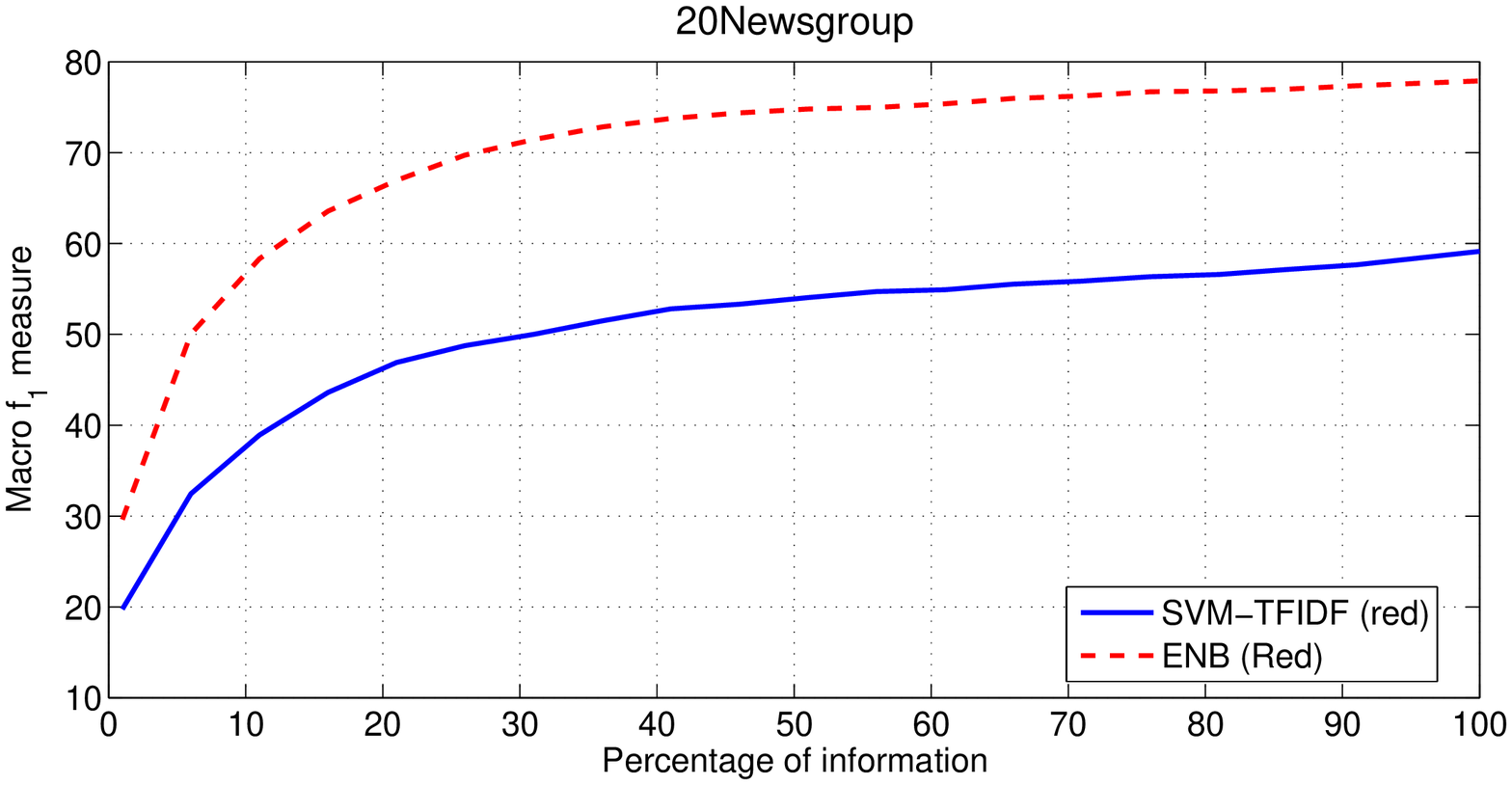}
    \includegraphics[scale=0.37]{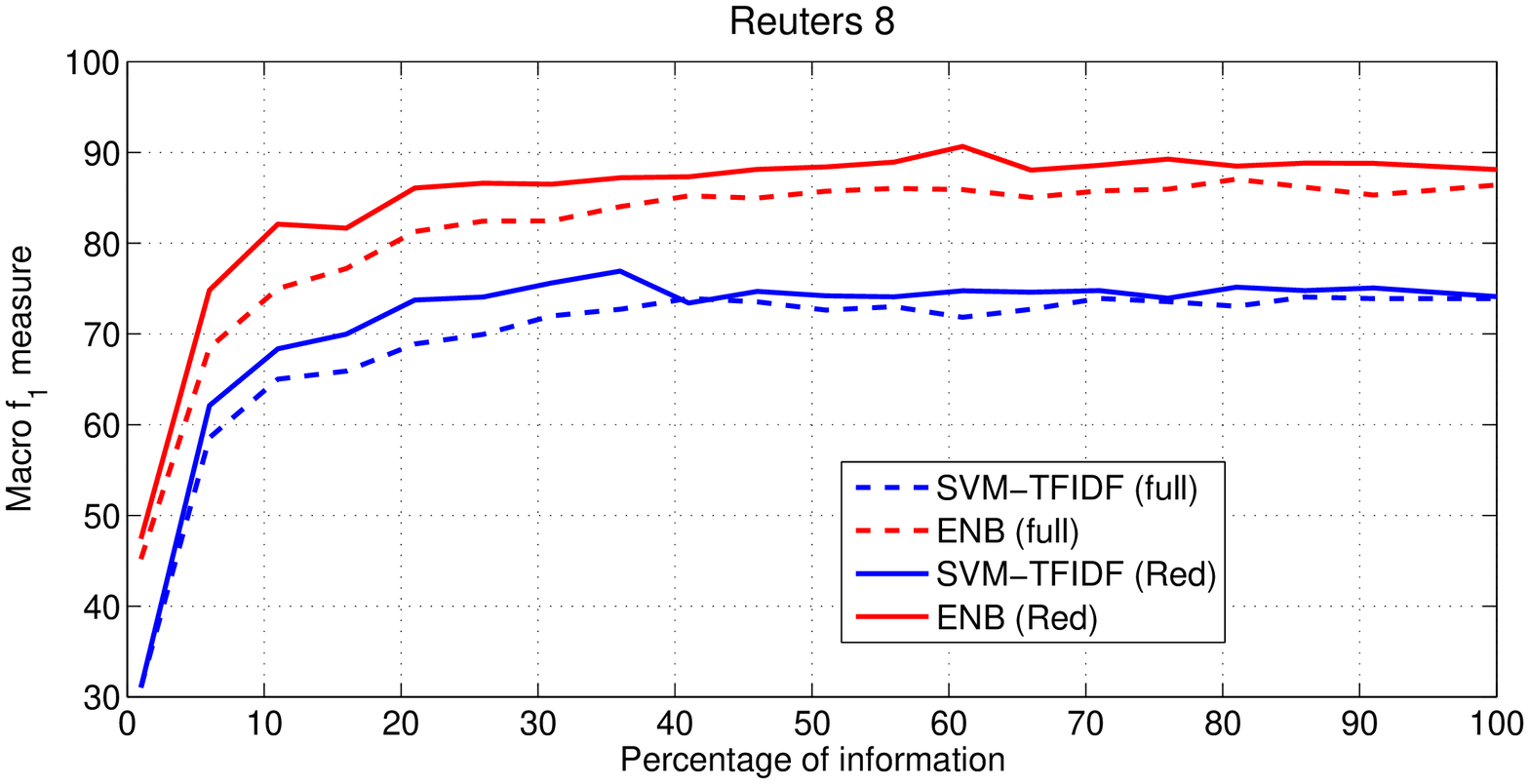}
    \includegraphics[scale=0.37]{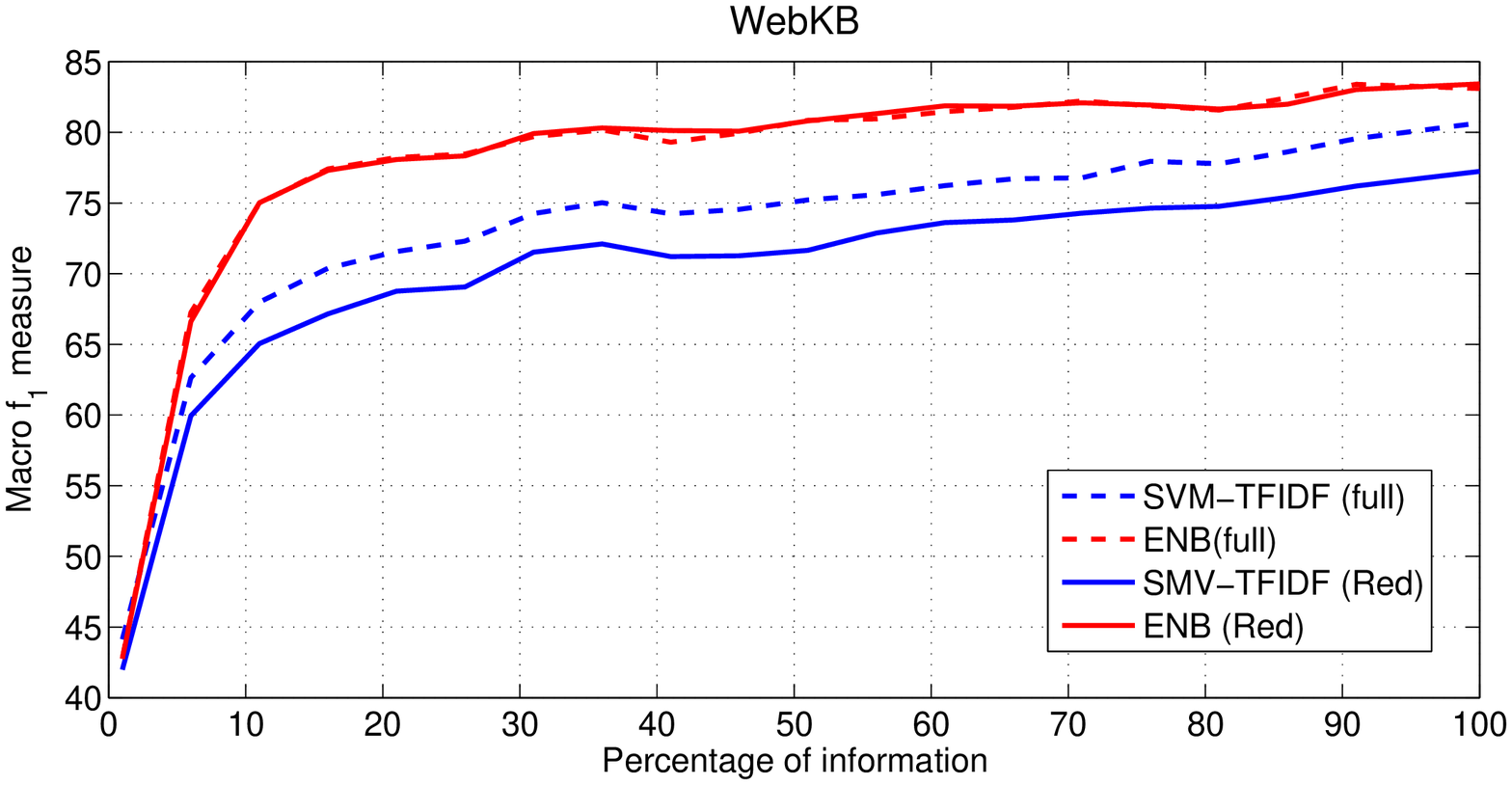}
    \caption{Early text classification on standard data sets. }\label{fig:20ngstandardp}
\end{figure}

First we analyze the performance of early na\"ive Bayes on thematic text classification. The first three data sets from Table~\ref{tab:datasets} were considered, these are widely used  benchmark data sets for text categorization; standard training/testing partitions\footnote{As reported in: http://web.ist.utl.pt/acardoso/datasets/} were used. Results of this experiment are shown in Figure~\ref{fig:20ngstandardp}. %,~\ref{fig:r8standardp} and~\ref{fig:webkbstandardp}. .
%\begin{figure}[htb!]
%    \centering
%    \includegraphics[scale=0.3]{r8_standard_partitions.eps}
%    \caption{Comparison of early na\"ive Bayes (ENB) and the reference method (STC).}\label{fig:r8standardp}
%\end{figure}
%\begin{figure}[htb!]
%    \centering
%    \includegraphics[scale=0.3]{webkb_standard_partitions.eps}
%    \caption{Comparison of early na\"ive Bayes (ENB) and the reference method (STC).}\label{fig:webkbstandardp}
%\end{figure}

It can be seen in the top plot that the early na\"ive Bayes (ENB hereafter) classifier outperforms considerably the SVM baseline for the 20Newsgrup data set. For both methods, the performance increased monotonically and, as  expected, better performance was obtained when more information is considered. 

The middle and bottom plots in Figure~\ref{fig:20ngstandardp} show results for Reuters 8 and WebKB, respectively; in these plots we show the performance of both methods, ENB and SVM, and when using all of the vocabulary (\emph{full}) and a reduced one (for 20Newsgrup data set we were not able to run an experiment with the full vocabulary in reasonable times). Regardless of the vocabulary used, ENB outperforms SVM. However, using the full vocabulary had opposed effects in the two data sets. In Reuters 8, using the whole vocabulary reduced the performance of both methods mainly when using less than $50\%$ of information; in WebKB the performance of ENB is virtually the same, but the performance of SVM increased when using the full vocabulary. This can be due to the specific characteristics of the data. Finally, in the three data sets it is somewhat evident that the predictive performance of ENB presents low variations after processing about $50\%$ of the texts.

\subsection{Comparison with related work}
\begin{figure}[htb!]
    \centering
    \includegraphics[scale=0.38]{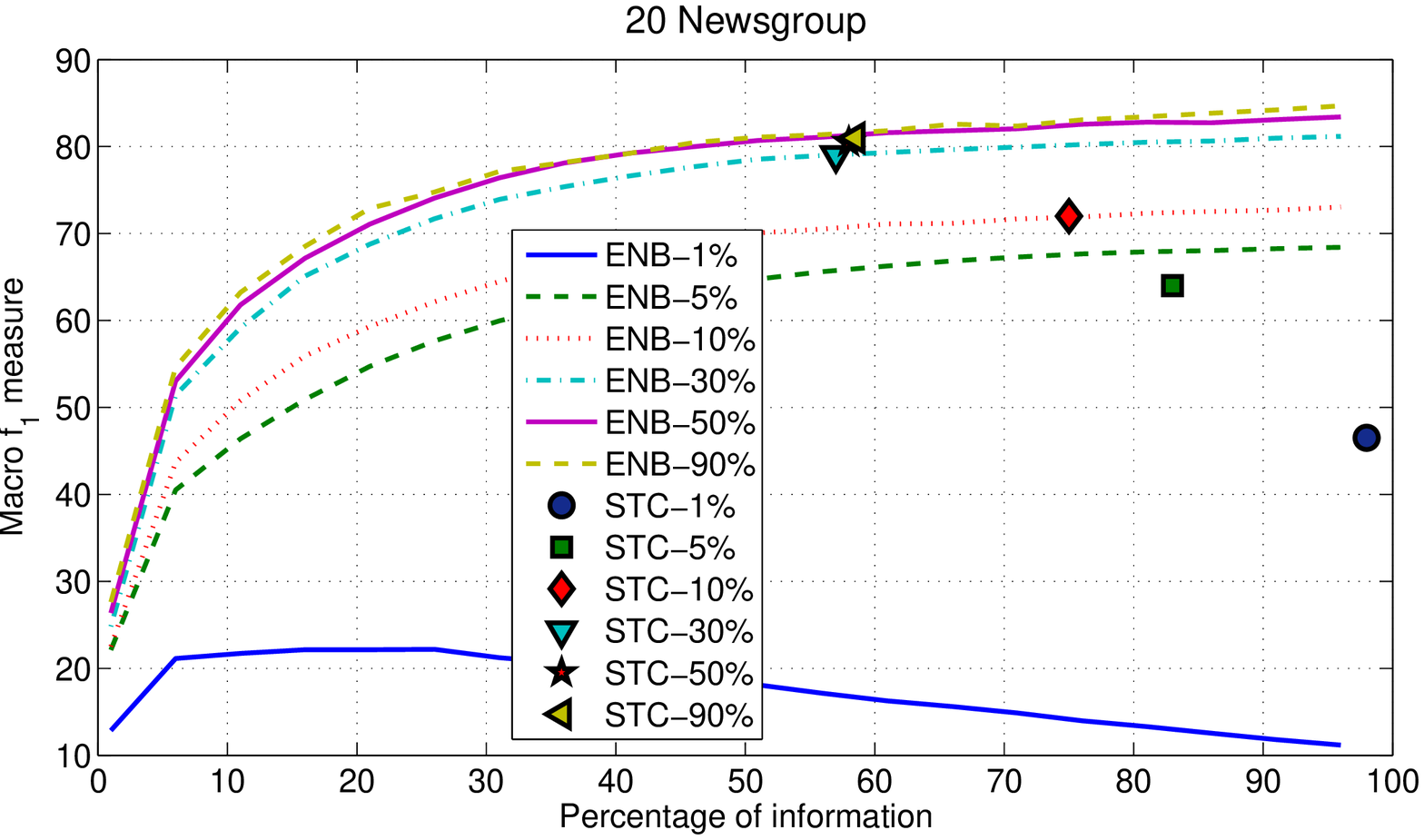}
    \includegraphics[scale=0.38]{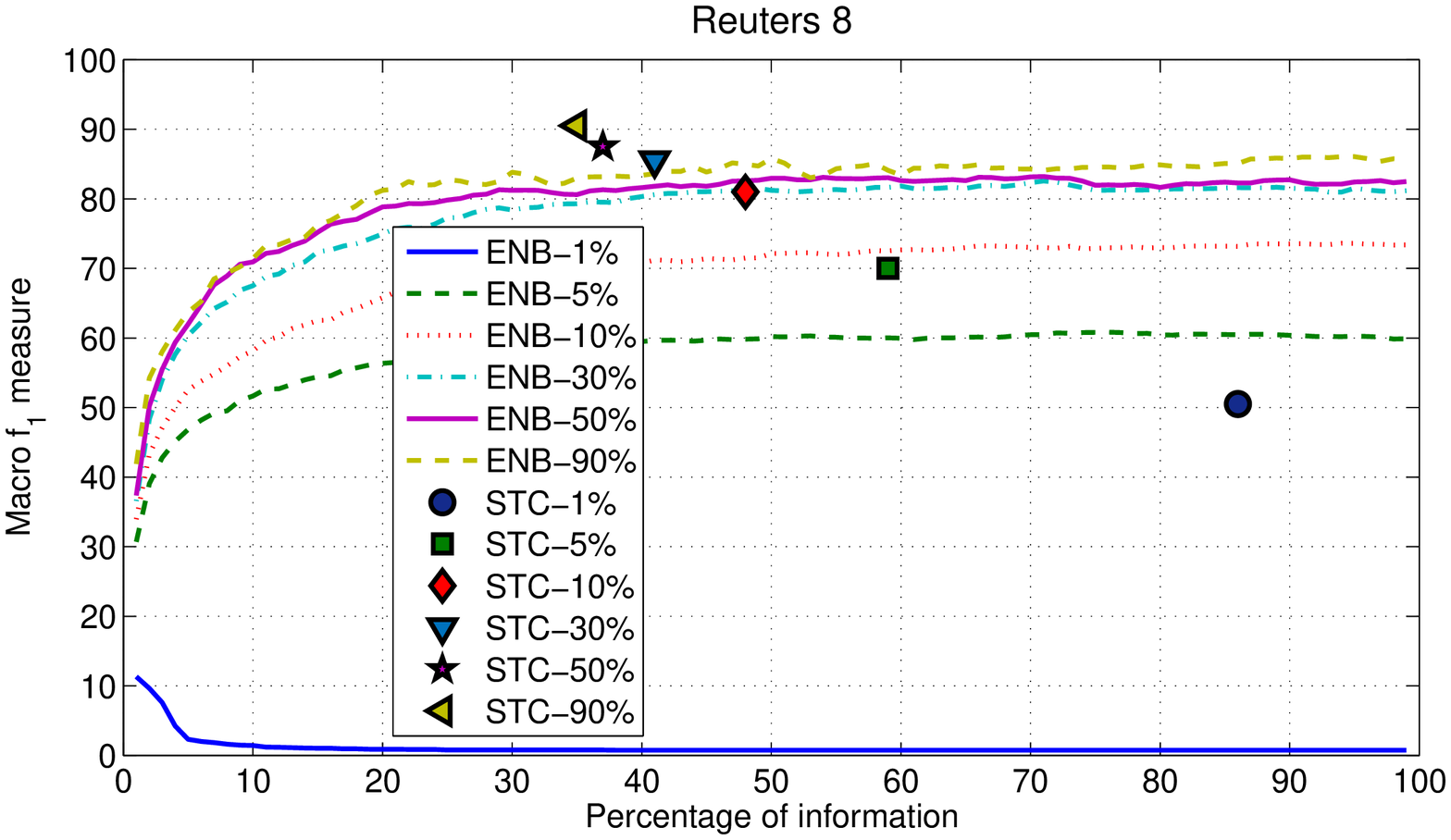}
    \includegraphics[scale=0.38]{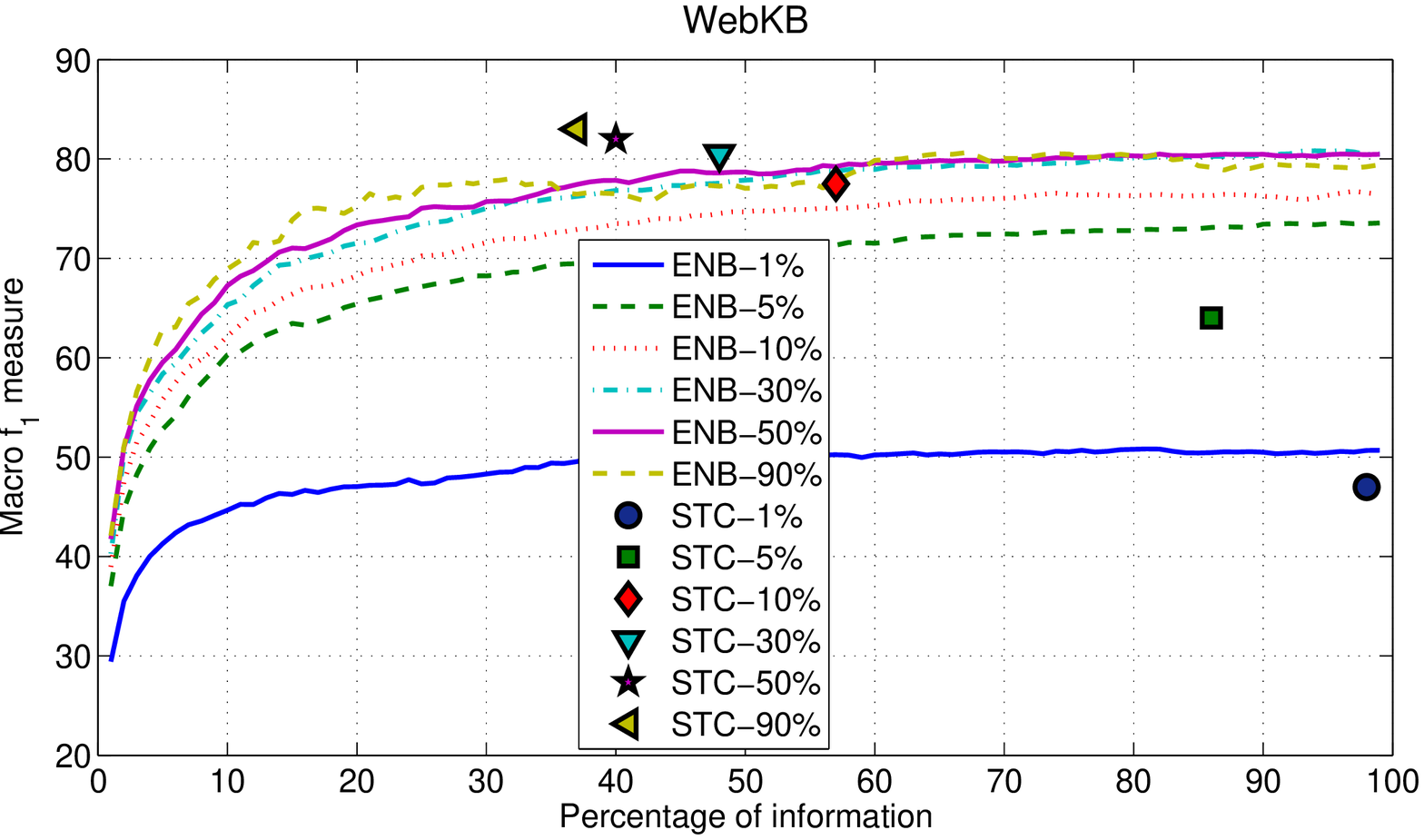}
    \caption{Comparison of ENB and the reference method STC.}\label{fig:comparisonr8}
\end{figure}

In this section we compare the performance of na\"ive Bayes with the MDP introduced in~\cite{sequentialTC} using the same data sets from the previous section. For this comparison we replicated the experiment reported by the authors of~\cite{sequentialTC}. For each of the data sets, we used different percentages, $\{1\%, 5\%, 10\%, 30\%, 50\%, 90\%\}$, of documents for the training set and the remainder for the test set (this was not our choice, but the setting proposed by the authors of the reference paper). Five runs were performed, in each run the documents for training were randomly chosen. Average results are shown in Figure~\ref{fig:comparisonr8}. 
The results of ENB are shown as graphs, whereas for the reference method we report the single-best reported result (shown as markers, one per training set size). Please note that in~\cite{sequentialTC} the authors optimized the parameters of their method, called STC, whereas we have used default implementation/parameters for ENB.

From Figure~\ref{fig:comparisonr8}, it can be seen that the percentage of training documents used for learning the model affects considerably the performance of ENB. In all three cases, using less than $30\%$ of the samples for training results in low performance. This can be due to the fact that with small amounts of training documents, the estimated probabilities are not very representative of the classification task (and so, it is not convenient to estimate probabilities from partial information only).
The best results were obtained when using $50\%$ or $90\%$  of instances for training the model. Also we can notice that the performance stabilizes after $40\%$ of the information has been processed. 

When comparing  the ENB approach with the sequential text classification technique (STC) from~\cite{sequentialTC}, it can be seen that %the performance of 
the MDP from the reference work and our ENB perform very similar (even when we only show best/optimized results for STC). This is a very interesting result: we obtained comparable performance to a more complex model, with a much more simpler and efficient technique.

\subsection{Sexual predator detection}

\begin{figure}[htb!]
    \centering
    \includegraphics[scale=0.36]{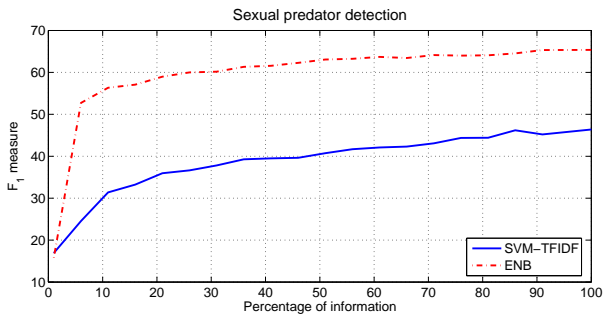}
    \caption{Early classification performance on detection of sexual predators. }\label{fig:sexpred}
\end{figure}

We now evaluate the performance of ENB on the task of sexual predator detection. We used the development / test partitions  of the data set used in the sexual predator competition from PAN'12~\cite{sexpred}, see Table~\ref{tab:datasets}. This corpus contains a large number of chat conversations, some of which include a sexual predator trying to approach a child\footnote{Police officers acted as children, predators are real. }. The problem approached in the original competition was to identify sexual predators from many chat conversations. However, in this work, we approach the problem of detecting conversations with potential sexual predators in it.
We proceeded in this way because the original task was one of forensic analysis: detect predators offline using all of the conversations in which they were involved (see~\cite{villatoro} for our solution that obtained the best result in that challenge). Our ultimate goal, on the other hand, is to detect, as early as possible, conversations in which a sexual predator is involved, in such a way that sexual-attacks can be prevented and an alert for parents/police officers can be emitted.
%develop preventive mechanisms for protecting children. % (a tool for triggering alerts for parents would be very helpful).
%Clearly, the approached problem demands early classification models, as one wants to detect the predator before he hurts  a child (i.e., prevention) and not after (forensics). We approach the problem of classifying conversations as including a sexual predator or not, a reduced vocabulary was used.
Based on our previous results from~\cite{villatoro}, and on the literature on non-thematic text classification we decided to represent chat conversations with 3-grams of characters (i.e., terms in this data set are sequences of 3-letters extracted from the training corpus); with this data set we used a reduced vocabulary and preprocessing  processes described in~\cite{villatoro}.
%we used a reduced vocabulary, as suggested in the literature, we used character 3-grams to represent chat conversations~\cite{sexpred}. 
As suggested in~\cite{sexpred}, for this experiment we report $f_1$ measure on the minority class (i.e., predators). Results of this experiment are shown in Figure~\ref{fig:sexpred}.

On the one hand, we can see that this is a very difficult task, the performance of both models, SVM and ENB, is somewhat low, even when the whole information from documents is used (the highest performance is lower than 70\% of $f_1$ measure). This is not a surprising result if we notice that this problem is highly imbalanced: the imbalance ratio for training and test partitions is of $12.1$ and $9.56$, respectively. Furthermore, the reduction of the vocabulary may affect significantly this particular domain (the jargon used in chat conversations is quite diverse and rich).  Despite the difficulty of the problem, we can see that again the ENB method outperforms the SVM model in most cases. %The behaviour of ENB is monotonic increasing, whereas that of SVM is somewhat erratic, in some cases it outperforms the ENB technique.
Results shown in this section make evident the need of better methods for early text classification.

\section{Conclusions}
\label{sec:conclusions}
We described the use of na\"ive Bayes for early text classification. A minor modification to na\"ive Bayes allows us to make predictions using partial information. We show the effectiveness of this simple approach in three types of problems %(one involving video) 
and compare its performance with the only existing state-of-the-art method. Our method compares favorably in terms of both effectiveness and earliness performance with the reference method, a much more complex model. Also, our method consistently outperformed an SVM baseline. Furthermore, we are the first in approaching the early classification of chat conversations for detecting sexual predators. Although results are encouraging, there is too much work to do yet.  We foresee our work will pave the way for the development of more elaborated techniques based on na\"ive Bayes for early classification. 

The following conclusions can be drawn from our work:
\begin{itemize}
\item Na\"ive Bayes proved to be very effective for early text classification, obtaining comparable results to state of the art. The inference complexity of na\"ive Bayes is negligible (adding the value of $q-$terms, for $K-$times), thus makes this method preferable over the MDP introduced in~\cite{sequentialTC}.
\item Na\"ive Bayes is a promising solution to the early classification problem. Competitive performance was obtained with a somewhat straight implementation, better results are expected with improved versions of the classifier. 
\item It is possible to anticipate the detection of sexual predators, being na\"ive Bayes a potential  solution to this problem. 
\end{itemize}

Future work is vast, for instance, exploiting research advances in extensions of na\"ive Bayes (see Section~\ref{sec:rw}) for early text classification. Also, it is very important to develop spotting mechanisms that can be combined with the early na\"ive Bayes technique. Finally, theoretical analyses of the problem and the proposed method
are very much needed. %\subsection{Length of Papers}
%

\bibliographystyle{plain}
\bibliography{sdm}

\end{document}